\documentclass[10pt,twocolumn,letterpaper]{article}

\usepackage{cvpr}              

\usepackage{amsmath}

\usepackage{bbding}
\usepackage{pifont}
\usepackage{wasysym}
\usepackage{amssymb}

\usepackage{verbatim}
\usepackage{booktabs}
\usepackage{multirow}
\usepackage{colortbl}
\usepackage{adjustbox}
\usepackage{makecell}
\usepackage{xspace}
\usepackage{subcaption}

\usepackage{color}    
\usepackage{enumitem}

\usepackage{times}
\usepackage{soul}
\usepackage{url}
\usepackage{bbding}
\usepackage{pifont}
\usepackage{wasysym}
\usepackage{amssymb}

\usepackage{verbatim}
\usepackage{booktabs}
\usepackage{multirow}
\usepackage{colortbl}
\usepackage{adjustbox}
\usepackage{makecell}
\usepackage{xspace}
\usepackage{booktabs}
\usepackage{multirow}
\usepackage{graphicx}
\usepackage{caption}
\usepackage{float}
\usepackage{bm}
\usepackage{makecell}

\definecolor{cvprblue}{rgb}{0.21,0.49,0.74}
\usepackage[pagebackref,breaklinks,colorlinks,allcolors=cvprblue]{hyperref}

\def\confName{CVPR}
\def\confYear{2026}

\title{Tempo as the Stable Cue: Hierarchical Mixture of Tempo and Beat Experts for Music to 3D Dance Generation}

\author{Guangtao Lyu$^{1}$, Chenghao Xu$^{2}$, Qi Liu$^{1}$, Jiexi Yan$^{3}$, Muli Yang$^{4}$, Fen Fang$^{4}$,  Cheng Deng$^{1}$\thanks{Corresponding author} \\
        $^{1}$ School of Electronic Engineering, Xidian University, China,   $^{2}$ Hohai university, China, \\  
        $^{3}$ School of Computer Science and Technology, Xidian University, China, \\ 
        $^{4}$ Institute for Infocomm Research (I\textsuperscript{2}R), A*STAR, Singapore,\\  
        \texttt{ \{guangtaolyu,qiliu\}@stu.xidian.edu.cn, fang fen@a-star.edu.sg},\\ \texttt{\{jxyan1995,muliyang.xd,chdeng.xd\}@gmail.com}}

\begin{document}
\maketitle
\begin{abstract}
Music to 3D dance generation aims to synthesize realistic and rhythmically synchronized human dance from music.
While existing methods often rely on additional genre labels to further improve dance generation, such labels are typically noisy, coarse, unavailable, or insufficient to capture the diversity of real-world music, which can result in rhythm misalignment or stylistic drift.
In contrast, we observe that tempo, a core property reflecting musical rhythm and pace, remains relatively consistent across datasets and genres, typically ranging from 60 to 200 BPM.
Based on this finding, we propose TempoMoE, a hierarchical tempo-aware Mixture-of-Experts module that enhances the diffusion model and its rhythm perception. TempoMoE organizes motion experts into tempo-structured groups for different tempo ranges, with multi-scale beat experts capturing fine- and long-range rhythmic dynamics. A Hierarchical Rhythm-Adaptive Routing dynamically selects and fuses experts from music features, enabling flexible, rhythm-aligned generation without manual genre labels.
Extensive experiments demonstrate that TempoMoE achieves state-of-the-art results in dance quality and rhythm alignment.
\end{abstract}

\section{Introduction}

Music to 3D dance generation aims to automatically synthesize realistic, diverse, and rhythmically aligned human dance conditioned on a given music sequence.  This task is fundamental for applications in virtual humans, digital choreography, and audio-visual content creation~\cite{dance_m2d_2006,dance_m2d_2011}, and also offers insights into the neural coupling of auditory and motor cognition~\cite{dance_neuroscience_old}.

Recent advances in deep generative modeling, spanning generative adversarial networks (GANs)\cite{m2d_dancing_to_music_dance_gan,m2d_music2dance}, autoregressive models~\cite{m2d_bailando,m2d_aistaichoreographer_fact_aist++,m2d_ar_diffusion_badm}, and diffusion models~\cite{m2d_beatit,m2d_align,m2d_popdg_popdanceset,m2d_edge,m2d_finedance}, have greatly improved the realism, smoothness, and diversity of synthesized dances.

Beyond improving generative architectures, recent works also leverage auxiliary information to enhance dance generation quality, most commonly through genre conditioning.
Early approaches represent dance styles using one-hot genre vectors~\cite{m2d_music2dance,m2d_dancing_to_music_dance_gan}; subsequent works learn continuous genre embeddings to capture richer stylistic variations~\cite{m2d_bailando,m2d_finedance}; more recent studies employ natural language prompts for flexible genre and style control~\cite{m2d_prompt,m2d_prompt_dancemosaic}.

\begin{figure}
    \centering
    \includegraphics[width=1\linewidth]{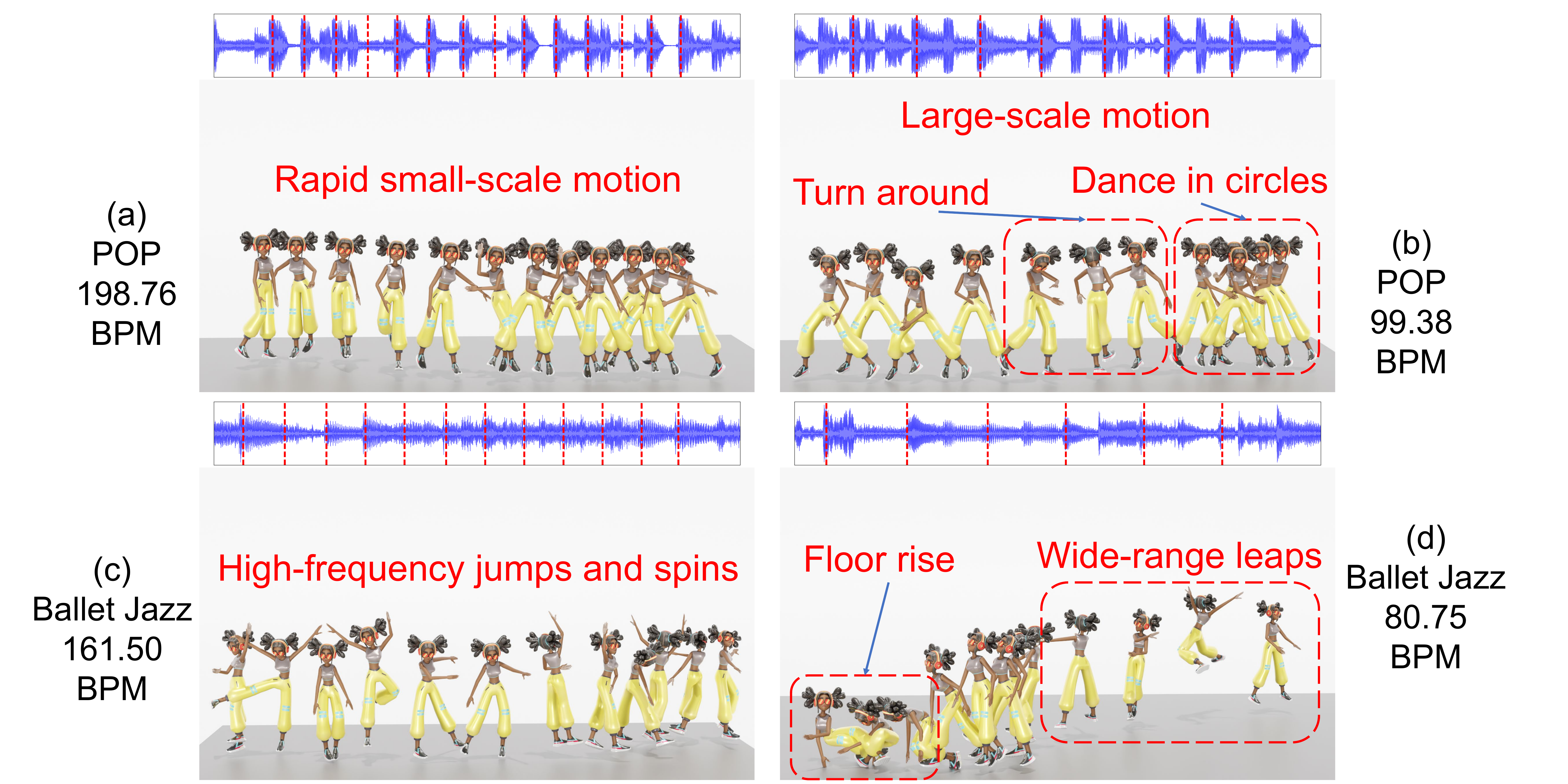}
\vspace{-20pt}
\caption{Visualization of dances under different tempos within the same genre. Even within a single genre, varying BPMs lead to distinct motion patterns: high BPM gives less time per beat, resulting in faster, more localized motions (e.g., quick arm swings, spins), while low BPM allows more time, supporting longer and more complex gestures (e.g., body turns, full-body transitions).}
\vspace{-16pt}
\label{fig:motivation_different_tempo}
\end{figure}

\begin{figure*}[t]
    \centering
    \includegraphics[width=1\linewidth]{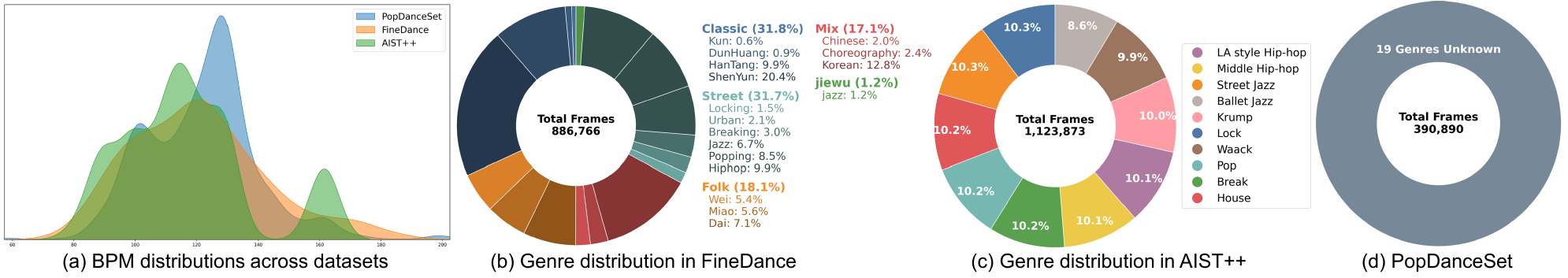}
    \vspace{-20pt}
\caption{ (a) BPM distributions across AIST++, FineDance, and PopDanceSet indicate that musical tempos predominantly fall within a shared range of 60–200 BPM, reflecting a common underlying rhythmic structure.   (b–d) In contrast, genre distributions are highly imbalanced and dataset-specific: FineDance and AIST++ adopt distinct genre taxonomies, while PopDanceSet provides no explicit genre annotations.   These insights motivate us to leverage BPM as a more reliable cue than genre labels. }
    \vspace{-12pt}
    \label{fig:motivation}
\end{figure*}

However, using one-hot genre labels as auxiliary cues for dance generation faces several limitations: 
(1) {Coarse Rhythmic Granularity}: dances within the same genre can exhibit widely varying tempo (BPM), intensity, and motion style. High BPM results in faster, more compact movements, while low BPM allows slower, broader, and more intricate gestures. A single genre token is therefore insufficient to capture such intra-genre rhythmic diversity~\cite{m2d_aistaichoreographer_fact_aist++}; see also Figure~\ref{fig:motivation_different_tempo}. 
(2) {Limited Scalability}: music and dance continuously evolve, giving rise to new styles, subgenres, and hybrids~\cite{dance_development_21st,dance_development_discover_new,dance_development_history_new}. Fixed genre tokens cannot accommodate this open-ended and dynamic style space. 
(3) {Unreliable Supervision}: real-world datasets often contain sparse, ambiguous, or inconsistent genre annotations. For example, PopDanceSet~\cite{m2d_popdg_popdanceset} lacks genre labels entirely (Figure~\ref{fig:motivation}(d)), making genre-conditioned generation unreliable. 
These issues limit a model’s ability to accurately capture rhythm and produce high-quality, diverse dance motions.

These limitations motivate us to explore alternative cues that can reliably enhance rhythmic dance generation. 
We find that tempo, a core property reflecting musical pace and typically measured in BPM, remains stable across datasets and genres~\cite{dance_tempo_1,dance_tempo_2,dance_tempo_4_what_is_musical_tempo}. 
As shown in Figure~\ref{fig:motivation}, despite the diversity of dance styles, BPM cluster within a consistent range (60–200 BPM), revealing a universal rhythmic structure that can serve as a stable prior to enhance the generation model's perception of rhythm and beat.

Motivated by these observations, we exploit the stable tempo cue to improve the diffusion model’s architecture and strengthen its rhythm-aware motion modeling. 
However, generating tempo-aware motions remains challenging due to the broad and continuous distribution of tempos (Figure~\ref{fig:motivation}). 
High-tempo requires rapid and intricate movements, demanding precise modeling of short-term, fine-grained dynamics, whereas low-tempo typically features smooth, continuous gestures evolving over longer durations (Figure~\ref{fig:motivation_different_tempo}), necessitating modeling of global motion structure~\cite{dance_tempo_3,dance_tempo_5_tempo_music_dance_synchronization}. 
Therefore, the generation model struggles to capture diverse tempo characteristics simultaneously, often leading to degraded motion quality or rhythm inconsistencies.

To effectively decouple these rhythmic variations, we seek an effective architecture that can adaptively allocate modeling capacity across different tempo conditions. To this end, Mixture-of-Experts (MoE)~\cite{moe_first_adaptive,moe_first_hier} offers a natural solution. MoE decomposes the model into multiple specialized sub-networks called experts, each capable of handling specific input patterns. A learned routing network dynamically selects and combines experts conditioned on input features, allowing the model to scale capacity efficiently and specialize experts across conditions. This clear separation between routing and expert processing has proven effective in vision, speech, and multimodal tasks~\cite{moe_moellava,moe_unimoe}.

Building on these insights, we introduce \textbf{TempoMoE}, a tempo-aware MOE framework that incorporates tempo and beat priors into the diffusion model, allowing experts to specialize in distinct BPM ranges and generate motions that are rhythmically aligned and structurally coherent.
Concretely, TempoMoE comprises two key components. The first is the Tempo-Structured Expert Groups. We discretize the tempo range into eight 20 BPM bands, as 20 BPM aligns with human perceptual thresholds and common dance practices reported in prior studies~\cite{dance_tempo_6_disco_time_split_20,dance_tempo_7_class}. Each tempo band corresponds to a dedicated expert group, whose temporal receptive field is designed to align with the motion pacing at that tempo level. Within each group, we define three specialized experts, each operating at a distinct beat scale: quarter-beat, half-beat, and full-beat, capturing rhythmic structures from fine-grained accents to phrase-level transitions.

The second key component is a two-stage routing mechanism, termed Hierarchical Rhythm-Adaptive Routing. In the first stage, we perform Hard Tempo-Level Group Selection using a lightweight gating network, TempoGateNet, which predicts the most relevant tempo groups based on the global music representation. We activate the top-2 groups to allow for flexible blending across tempo boundaries. In the second stage, we apply Soft Beat-Scale Expert Routing using BeatGateNet, which assigns soft gating weights to the three intra-group experts, enabling adaptive fusion across beat scales.
This hierarchical routing strategy allows the model to dynamically specialize along both tempo and rhythmic resolution dimensions, without relying on genre-specific annotations or fixed genre labels, leading to more accurate and expressive rhythm-aligned dance generation.

In summary, our main contributions are as follows:
\begin{itemize}
    \item We identify tempo as a stable auxiliary cue for improving dance generation, remaining consistent across datasets and genres without requiring manual annotation.

    \item We propose TempoMOE, which integrates tempo and beat priors into the diffusion model structure to enhance rhythmic synchronization and dance coherence.

    \item We introduce Tempo-Structured Expert Groups with multi-scale beat modeling, combined with a Hierarchical Rhythm-Guided Routing mechanism, enabling the model to adaptively generate motion dynamics across diverse tempos and rhythmic complexities.
    
    \item We conduct extensive experiments on AIST++, PopDanceSet, and FineDance, demonstrating SOTA performance in dance motion quality and rhythm alignment.

\end{itemize}

\begin{figure*}[t]
 \centering
 \includegraphics[width=1.0\textwidth]{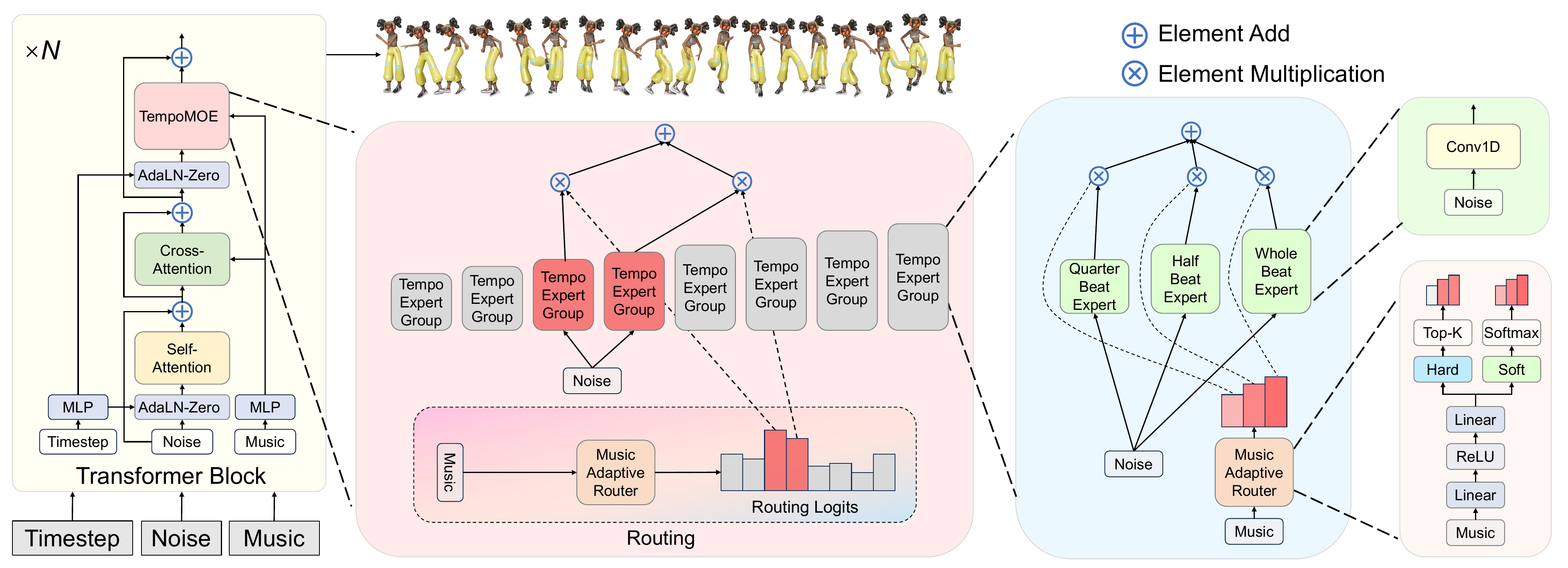}
 \vspace{-12pt}
\caption{Framework overview of TempoMoE, our tempo-aware dance generation diffusion model with $N$ transformer blocks. In addition to fusing music features via cross-attention, we replace the original FFN with TempoMoE, which adaptively activates tempo-specific expert groups containing multi-scale beat experts to synthesize coherent and rhythm-aligned 3D dance motion.}
 \vspace{-8pt}
 \label{fig:architecture}
\end{figure*}

\section{Related Work}\label{sec:related-work}

\textbf{Mixture of Experts.}
MoE~\cite{moe_first_hier,moe_first_adaptive} is a conditional computation framework that routes each input through a subset of specialized sub-networks (experts), offering two primary benefits:
(1) Capacity scaling without proportional compute, enabling ultra-large models~\cite{moe_large_sparse_gshard,moe_large_sparse_shared_expert_deepseekmoe}
(2) Dynamic expert specialization, where the gating function adaptively selects experts suited to each input, a strategy widely adopted across vision, speech, and multimodal modeling~\cite{moe_3dmoe,moe_unimoe,moe_moellava}. Routing in MoE is typically categorized into two types:
Hard routing activates the top-$k$ scoring experts and masks out the rest, often using Top-1 or Top-2 selection to balance efficiency and specialization~\cite{moe_large_sparse,moe_large_sparse_switchtransformers}.
Soft routing computes a weighted combination over all experts based on the gating distribution~\cite{moe_softmoe,moe_softmoe_smear}.

\noindent\textbf{Music-Driven Dance Generation.}
The task aims to produce temporally coherent, style-consistent dance motions conditioned on music~\cite{m2d_beatit,m2d_reward_rank_E3D2,m2d_logde++,m2d_first_li2020learning,m2d_choreomaster,m2d_long_dance_DanceRevolution,tmr_lexical_guangtao_ours,chenghao_guangtao_llm_acl,m2d_prompt_dancechat,m2d_choreograph}. Early methods relied on motion graphs and retrieval-based techniques~\cite{m2d_learn2dance,m2d_choreonet}, but suffered from poor rhythmic alignment and limited generalization. With the rise of deep learning, CNNs, RNNs, and GCNs were used to directly regress motions from music~\cite{m2d_dance_melody_lstm,m2d_gcn,m2d_music2dance}, followed by GAN-based models that improved realism and style fidelity~\cite{m2d_dancing_to_music_dance_gan}. Recent advances focus on generative modeling: autoregressive approaches~\cite{m2d_bailandopp,m2d_ar_diffusion_badm,m2d_align} enhance temporal consistency using learned music-style embeddings, while some diffusion models~\cite{m2d_prompt,m2d_prompt_dancemosaic} further improve synthesis quality via natural language or genre prompts.

The concurrent work MEGADance~\cite{m2d_megadance_moe} employs a MoE framework with genre-based gating to support multi-style dance generation. However, as discussed in the introduction, genre-conditioned modeling suffers from coarse rhythmic granularity, limited scalability, and unreliable supervision.
In contrast, we employ a tempo-informed MoE framework that routes based on tempo signals and leverages experts with diverse temporal scopes and priors, enabling diverse rhythmic modeling across varied styles.

\section{Preliminaries}

\noindent \textbf{Problem Definition.}
Music-to-dance generation aims to synthesize diverse and expressive 3D human motion sequences that are rhythmically aligned with an input music signal. Following prior works~\cite{m2d_long_dance_lodge}, we extract frame-level music features $c \in \mathbb{R}^{L \times 35}$ using Librosa~\cite{music_librosa}, capturing energy envelope, MFCCs, chroma, onsets, and beats, and represent motion sequences $\mathbf{x} \in \mathbb{R}^{L \times d}$ using standard conventions, including 3D root translation, 6D joint rotations, and binary foot contact signals.

\noindent \textbf{Diffusion Models.}
Diffusion models consist of two core components: a forward noising process and a reverse denoising process. Let $\mathbf{x}_0 \sim p(\mathbf{x}_0)$ denote a clean motion sample. The forward process gradually corrupts $\mathbf{x}_0$ by adding Gaussian noise over $T$ timesteps, resulting in a sequence $\{\mathbf{x}_1, \mathbf{x}_2, \ldots, \mathbf{x}_T\}$. This process is defined as:
\begin{align}
q(\mathbf{x}_t \mid \mathbf{x}_0) &= \mathcal{N}\big(\mathbf{x}_t;\, \alpha_t \mathbf{x}_0, \sigma_t^2 \mathbf{I} \big), \\
\mathbf{x}_t &= \alpha_t \mathbf{x}_0 + \sigma_t \boldsymbol{\epsilon}, \quad \boldsymbol{\epsilon} \sim \mathcal{N}(\mathbf{0}, \mathbf{I}),
\label{eq:diffusion_forward}
\end{align}
where $\alpha_t = \sqrt{1 - \sigma_t^2}$ and the noise scale $\sigma_t$ increases with timestep $t$. 
The reverse process aims to recover the original sample $\mathbf{x}_0$ from noisy inputs. It is modeled as a Gaussian transition:
\begin{equation}
p_\theta(\mathbf{x}_{t-1} \mid \mathbf{x}_t) = \mathcal{N}\big(\mathbf{x}_{t-1};\, \hat{\boldsymbol{\mu}}_\theta(\mathbf{x}_t, t), \hat{\boldsymbol{\Sigma}}_\theta(\mathbf{x}_t, t)\big),
\end{equation}
where $\hat{\boldsymbol{\mu}}_\theta$ and $\hat{\boldsymbol{\Sigma}}_\theta$ are estimated by a neural network. 

To train the denoising model, we adopt the clean sample prediction objective~\cite{diffusion_target_sample} for its compatibility with downstream constraints, following the previous methods~\cite{t2m_mdm,m2d_edge}:
\begin{equation}
\mathcal{L}_{\text{sample}} = \mathbb{E}_{\mathbf{x}_0, t} \left[ \left\| \mathbf{x}_0 - \hat{\mathbf{x}}_{\theta}(\mathbf{z}_t, t, \mathbf{c}) \right\|_2^2 \right],
\end{equation}
where $\mathbf{z}_t$ is the noisy input at timestep $t$ and $\mathbf{c}$ denotes the conditioning signal.

\section{Method}
\label{sec:method}

\subsection{Framework Overview}
We adopt a denoising diffusion transformer (DIT)~\cite{diffusion_dit} architecture for motion generation. Starting from a clean motion sequence $\mathbf{x}_0$, we iteratively add Gaussian noise over $T$ timesteps to produce noisy samples $\mathbf{x}_t$, and train the model to recover $\mathbf{x}_0$ through a learned reverse diffusion process. Each denoising block contains three components: (1) a self-attention module to model intra-motion dependencies; (2) a cross-attention module that integrates rhythm-aware music features $c$; and (3) an AdaLN-Zero~\cite{diffusion_dit} modulated feed-forward network for global conditioning.

To enhance rhythmic sensitivity and tempo adaptivity, we replace the feed-forward network with the Tempo-Aware Mixture-of-Experts (TempoMoE) module. TempoMoE dynamically routes features through a set of tempo-structured expert groups guided by a hierarchical rhythm-adaptive router. This design enables both precise beat-level modeling and flexible tempo generalization, improving alignment with musical structure across diverse tempos.

\subsection{Tempo-Structured Expert Groups}

Human dance movements often align with hierarchical rhythmic patterns in music, such as beat subdivisions and phrasing. To reflect this structure, we introduce Tempo-Structured Expert Groups that specialize in processing motion features at multiple tempo levels and beat scales.

We discretize the tempo range  into eight bands centered at anchor BPMs at intervals of 20 BPM~\cite{dance_tempo_6_disco_time_split_20}:
\begin{equation}
 \{60, 80, 100, 120, 140, 160, 180, 200\} \text{ BPM}.
\end{equation}
Each tempo band defines a Tempo-Level Expert Group comprising three Beat-Scale Experts, which operate at $\frac{1}{4}$-, $\frac{1}{2}$-, and 1-beat resolutions. These experts capture fine-grained beat accents and long-range phrasing, respectively.

Each expert is designed to capture motion at a specific beat scale and is implemented as a 1D temporal convolutional layer. By adjusting the kernel size, we control the temporal receptive field of each expert. Given the frame rate $f$ and tempo $\text{BPM}$, we compute the expected number of frames per beat to guide the kernel design:
\begin{equation}
 F_b = \frac{60 \times f}{\text{BPM}}, \quad k = \left\lceil r \times F_b \right\rceil_{\text{odd}},
\end{equation}
where $r$ is the beat subdivision (e.g., $\frac{1}{4}$) and $\left\lceil \cdot \right\rceil_{\text{odd}}$ ensures symmetric alignment.

In summary, Tempo-Structured Expert Groups decompose motion modeling across tempo levels and beat scales, allowing each expert to specialize in a specific temporal resolution. This hierarchical design captures both fine-grained beat-level accents and long-range phrasing, while smooth transitions between tempo bands enable robust adaptation to varying BPMs.

\subsection{Hierarchical Rhythm-Adaptive Routing}

To fully leverage the specialization of Tempo-Structured Expert Groups, we introduce a two-stage routing strategy called Hierarchical Rhythm-Adaptive Routing, which dynamically selects and fuses experts conditioned on both global tempo context and local rhythmic variation. It consists of two key components: TempoGateNet for tempo-level expert group selection and BeatGateNet for intra-group beat-scale routing.

\paragraph{Hard Tempo-Level Expert Group Routing.}  
To select the most relevant tempo-level expert groups, we first compute a score vector for all $G$ groups based on the music feature $\mathbf{c} \in \mathbb{R}^d$:
\begin{equation}
\mathbf{s} = \mathrm{TempoGateNet}(\mathbf{c}) \in \mathbb{R}^{G},
\end{equation}
where $s_g$ represents the activation score for the $g$-th tempo-level group. We then select the top-$K$ groups 
\begin{equation}
\mathcal{G}_{\text{sel}} = \text{TopK}(\mathbf{s}, K)
\end{equation}
for activation. This hard selection enforces computational efficiency and ensures that each active group specializes in its designated BPM range with a tailored temporal receptive field. Empirically, $K=2$ provides a good balance between flexibility and routing sharpness.

\paragraph{Soft Beat-Scale Expert Routing.}
Within each selected group $g \in \mathcal{G}_{\text{sel}}$, we apply BeatGateNet to softly attend to its three sub-experts (corresponding to $\frac{1}{4}$-, $\frac{1}{2}$-, and 1-beat resolutions). The attention weights are computed via:
\begin{equation}
\boldsymbol{\gamma}^{(g)} = \mathrm{Softmax}(\mathrm{BeatGateNet}(\mathbf{c})) \in \mathbb{R}^3,
\end{equation}
and the fused output for group $g$ is:
\begin{equation}
\mathbf{y}^{(g)} = \sum_{k=1}^{3} \gamma_k^{(g)} \cdot f_k^{(g)}(\mathbf{h}),
\end{equation}
where $f_k^{(g)}(\mathbf{h})$ denotes the output of the $k$-th beat-scale expert in group $g$ given motion features $\mathbf{h}$. Soft fusion across beat resolutions allows the model to dynamically interpolate between micro-beat details and macro-phrase dynamics, adapting to local rhythmic textures.

\paragraph{Final Fusion.}
The final output of the TempoMoE module is computed by aggregating the selected group outputs:
\begin{equation}
\mathbf{h}' = \sum_{g \in \mathcal{G}_{\text{sel}}} \mathbf{y}^{(g)}.
\end{equation}
This hierarchical design enables musically grounded control over both temporal span and rhythmic granularity.

\subsection{Training Objectives}

        
We jointly optimize diffusion reconstruction and kinematic consistency to enhance realism and rhythmic fidelity.

\textbf{Kinematic Loss.} To ensure physically plausible motion, we adopt a joint-level kinematic loss based on forward kinematics (FK), following prior work~\cite{m2d_edge,m2d_long_dance_lodge,m2d_popdg_popdanceset}. This loss penalizes unnatural joint rotations and enforces biomechanical constraints, improving motion realism and temporal coherence:
\begin{align}
 \mathbf{x}_j^i = FK(\mathbf{x}_0^i), & \quad
 \mathbf{x}_v^i = \text{Vel}(\mathbf{x}_0^i), \quad
 \mathbf{x}_a^i = \text{Acc}(\mathbf{x}_0^i), \\
 \mathcal{L}_{\text{joint}} &= \frac{1}{L} \sum_{i=1}^L \| \mathbf{x}_j^i - \hat{\mathbf{x}}_j^i \|_2^2, \\
 \mathcal{L}_{\text{vel}} &= \frac{1}{L} \sum_{i=1}^L \| \mathbf{x}_v^i - \hat{\mathbf{x}}_v^i \|_2^2, \\
 \mathcal{L}_{\text{acc}} &= \frac{1}{L} \sum_{i=1}^L \| \mathbf{x}_a^i - \hat{\mathbf{x}}_a^i \|_2^2, \\
 \mathcal{L}_{\text{contact}} &=\frac{1}{n-1} \sum_{j=1}^{n-1} \left\|  (\mathbf{x}_j^{i+1} - \mathbf{x}_j^i)  \cdot {\hat{b}^{(i)}} \right\|^2_2,
\end{align}
where $\hat{b}^{(i)}$ is the binary body contact label’s portion of the pose at each frame $i$.
Total kinematic loss:
\begin{equation}
 \mathcal{L}_{\text{kin}} = \lambda_{\text{joint}} \mathcal{L}_{\text{joint}} + \lambda_{\text{vel}} \mathcal{L}_{\text{vel}} + \lambda_{\text{contact}} \mathcal{L}_{\text{contact}} + \lambda_{\text{acc}} \mathcal{L}_{\text{acc}}.
\end{equation}

\textbf{Overall Loss.} The full training objective is:
\begin{equation}
 \mathcal{L}_{\text{total}} = \mathcal{L}_{\text{simple}} + \mathcal{L}_{\text{kin}}.
\end{equation}

\begin{table*}[t]
\vspace{-8pt}
\caption{Quantitative results on three datasets. Best and second-best results are marked in \textbf{bold} and \underline{underline}, respectively.}
\vspace{-8pt}
\label{tab:main_result}

\centering
\renewcommand\arraystretch{1.0}
\begin{tabular}{llcccccc}
\toprule
 & & \multicolumn{2}{c}{Motion Quality} & \multicolumn{2}{c}{Motion Divsersity} & Rhythmic & User Study \\ \cmidrule(r){2-2} \cmidrule(r){3-4} \cmidrule(r){5-6} \cmidrule(r){7-7} \cmidrule(r){8-8}
Dataset & Method & $\text{FID}_{k}$ $\downarrow$ & $\text{FID}_{g}$ $\downarrow$ & $\text{Div}_{k}$ $\uparrow$ & $\text{Div}_{g}$ $\uparrow$ & BAS $\uparrow$ & Ours Wins  \\
\midrule

\multirow{6}{*}{AIST++} 
& Ground Truth & -- & -- & 8.19 & 7.45 & 0.2374 & -- \\
\cmidrule{2-8}

& FACT & 35.35 & 22.11 & 5.94 & 6.18 & 0.2209 & 91.2\% \\
& Bailando & \underline{28.16} & \textbf{9.62} & \underline{7.83} & \underline{6.34} & 0.2332 & 81.6\% \\
& EDGE & 42.16 & 22.12 & 3.96 & 4.61 & 0.2334 & 81.2\% \\
& Lodge & 37.09 & 18.79 & 5.58 & 4.85 & \underline{0.2423} & 86.5\% \\
& TempoMOE & \textbf{25.13} & \underline{10.96} & \textbf{7.98} & \textbf{6.86} & \textbf{0.2446} & -- \\
\midrule

\multirow{8}{*}{FineDance} 
& Ground Truth & -- & -- & 9.73 & 7.44 & 0.2120 & -- \\
\cmidrule{2-8}
& FACT & 113.38 & 97.05 & 3.36 & 6.37 & 0.1831 & 93.4$\%$ \\
& Bailando & 82.81 & \underline{28.17} & 7.74 & 6.25 & 0.2029 & 81.5$\%$ \\
& EDGE & 94.34 & 50.38 & \underline{8.13} & \underline{6.45} & 0.2116 & 76.3$\%$ \\
& Lodge & \underline{50.00} & 35.52 & 5.67 & 4.96 & \underline{0.2269} & 69.1$\%$ \\
& TempoMOE & \textbf{38.42} & \textbf{25.62} & \textbf{8.57} & \textbf{6.83} & \textbf{0.2316} & -- \\
\midrule

\multirow{6}{*}{PopDanceSet} 

& Ground Truth & -- & -- & 8.32 & 7.68 & 0.2603 & -- \\
\cmidrule{2-8}
& FACT & 37.62 & 26.32 & 5.63 & 6.13 & 0.2162 & 86.3\% \\
& Bailando & 29.56 & 22.47 & 5.92 & 6.29 & 0.2253 & 82.3\% \\
& EDGE & 34.58 & 23.72 & 6.13 & \underline{6.48} & 0.2334 & 80.9\% \\
& POPDG & 27.13 & \underline{21.41} & \underline{6.52} & 6.37 & \underline{0.2403} & 78.4\% \\
& TempoMOE & \textbf{23.42} & \textbf{16.18} & \textbf{7.54} & \textbf{7.12} & \textbf{0.2482} & -- \\

\bottomrule
\end{tabular}

\end{table*}

\section{Experiments}\label{sec:experiments}

\paragraph{Datasets.}
 \textbf{AIST++}~\cite{m2d_aistaichoreographer_fact_aist++} contains 1,408 high-quality motion sequences across 10 street dance genres, with multiple tempos per genre. Many sequences feature identical choreographies performed to different BPMs.
 \textbf{FineDance}~\cite{m2d_finedance} features over 7.7 hours of professional dance data captured via optical MoCap, covering 22 genres. The average sequence length exceeds 150 seconds, supporting evaluation on long-range choreography.
 \textbf{PopDanceSet}~\cite{m2d_popdg_popdanceset} is an in-the-wild dataset collected from BiliBili, featuring 19 diverse styles without  genre labels. It serves as a challenging benchmark for evaluating youth-oriented dance generation.

\paragraph{Evaluation Metrics.}
Following prior works~\cite{m2d_bailando,m2d_popdg_popdanceset}, we use $\text{FID}_k$ and $\text{FID}_g$ to measure motion quality, $\text{DIV}_k$ and $\text{DIV}_g$ to assess motion diversity, and Beat Alignment Score (BAS) to evaluate rhythm synchronization.

\begin{table}[t]
\vspace{-8pt}
\caption{Comparison of model architecture and efficiency. Reported runtime includes only motion generation, excluding music feature extraction. When feature extraction is included, our efficiency advantage is even greater, since lightweight Librosa features are much faster to obtain than Jukebox embeddings, which require an additional model.}
\label{tab:model_efficiency}
\vspace{-8pt}
\centering
 \resizebox{0.98\linewidth}{!}{
\begin{tabular}{lcccccc}
\toprule
\textbf{Method} & \textbf{Music Feature} & \textbf{\#Model} & \textbf{Params (M)} & \textbf{Steps} & \textbf{Time (s)} \\
\midrule
EDGE   & Jukebox 4800 dim & Single & 49.91  & 50  & 1.3  \\
POPDG  & Jukebox 4800 dim & Single & 101.41  & 50  & 2.8  \\
Lodge  & Librosa 35 dim   & Two    & 108.23 & 100 & 0.8  \\
\textbf{Ours}  & Librosa 35 dim   & Single & 70.13  & 10  & 0.6 \\
\bottomrule
\end{tabular}}

\end{table}

\paragraph{Implementation Details.}
The diffusion model has 8 blocks, latent dim 512, and is trained for 500 epochs with Adam (lr 1e-4, 100-step warmup, batch 128). Class-free guidance uses 10 \% dropout in training and a scale of 2.5 at inference. Sampling employs 10-step DPM-Solver++~\cite{diffusion_dpm_solver_++}. Following prior works~\cite{m2d_edge}, loss weights are $\lambda_{\text{joint}}=0.646$, $\lambda_{\text{vel}}=2.964$, $\lambda_{\text{contact}}=10.942$, $\lambda_{\text{acc}}=1$. All experiments run on NVIDIA A6000 GPUs.

\subsection{Performance Comparison}

\paragraph{Quantitative Results.}
We evaluate our method on three benchmarks: AIST++, FineDance, and PopDanceSet. Table~\ref{tab:main_result} shows comparisons with existing methods, including Fact~\cite{m2d_aistaichoreographer_fact_aist++}, Bailando~\cite{m2d_bailando}, EDGE~\cite{m2d_edge}, and Lodge~\cite{m2d_long_dance_lodge}. Our method consistently outperforms these methods across all datasets and metrics. Notably, we achieve significant improvements in BAS, highlighting the strength of our tempo-aware expert routing. Higher diversity metrics (\text{DIV}$_k$, \text{DIV}$_g$) also indicate our model generates more varied and expressive dances without compromising coherence, validating its effectiveness under diverse tempos and styles.

\begin{table}[t]
\vspace{-8pt}
\caption{Ablation results on Expert Group. The results demonstrate the effectiveness of our heterogeneous multi-scale design in capturing diverse and rhythmically aligned motions.}
\vspace{-8pt}
\label{tab:abla_expert_group}
    \centering
\renewcommand\arraystretch{1.0}
\begin{tabular}{l|ccc}
     \toprule
     Method & ${\text{FID}_k} \downarrow$ & ${\text{Div}_k} \uparrow$  & $\text{BAS} \uparrow$ \\
     \midrule
     Homo. Same-Scale & 34.52 & 6.23 & 0.2126 \\
     Homo. Multi-Scale & 29.73 & 6.92 & 0.2249 \\
     Hetero. Multi-Scale (Ours) & \textbf{25.13} & \textbf{7.98} & \textbf{0.2446} \\
     \bottomrule
\end{tabular}

\end{table}

\begin{table}[t]
\vspace{-8pt}
\caption{Ablation results on Multi-Scale Beat Experts. The results validate the effectiveness of using multi-scale beat experts within each tempo-level group.}
\vspace{-8pt}
\label{tab:abla_expert_multiple_beat_scale_expert}
    \centering
\renewcommand\arraystretch{1.0}
\begin{tabular}{l|ccc}
     \toprule
     Method & ${\text{FID}_k} \downarrow$ & ${\text{Div}_k} \uparrow$  & $\text{BAS} \uparrow$ \\
     \midrule
     Quarter-Only & 34.52 & 6.16 & 0.2249 \\
     Half-Only & 31.62 & 6.42 & 0.2213 \\
     Whole-Only & 28.92 & 6.86 & 0.2142 \\
     Mixed (Ours) & \textbf{25.13} & \textbf{7.98} & \textbf{0.2446} \\
     \bottomrule
\end{tabular}

\end{table}

\paragraph{Qualitative Results.}
Figure~\ref{fig:vis_m2d} presents qualitative comparisons of dance generation across different music genres. Compared to existing methods, which often suffer from drifting, off-beat, or repetitive motions, our approach achieves improved temporal coherence, greater motion diversity, and more expressive, genre-adapted performances. The generated sequences are visually natural, stylistically consistent, and closely aligned with the audio beat structures. By effectively capturing tempo-dependent motion patterns and diverse stylistic nuances, these results demonstrate the effectiveness of our tempo-aware MoE design in producing rhythmically accurate and expressive dances.

\paragraph{User Study.}
To evaluate real-world user perception, we conducted a user study on 20 participants with dance experience. Each was presented with 20 video pairs, comparing our method against existing methods, and asked to choose the better dance based on visual quality, rhythm synchronization, and stylistic match with the music. As shown in Table~\ref{tab:main_result}, our method received consistently higher preferences, highlighting its ability to generate rhythmically compelling and user-favored dance in realistic settings.

\begin{figure*}[t]
    \centering
    \includegraphics[width=0.99\linewidth]{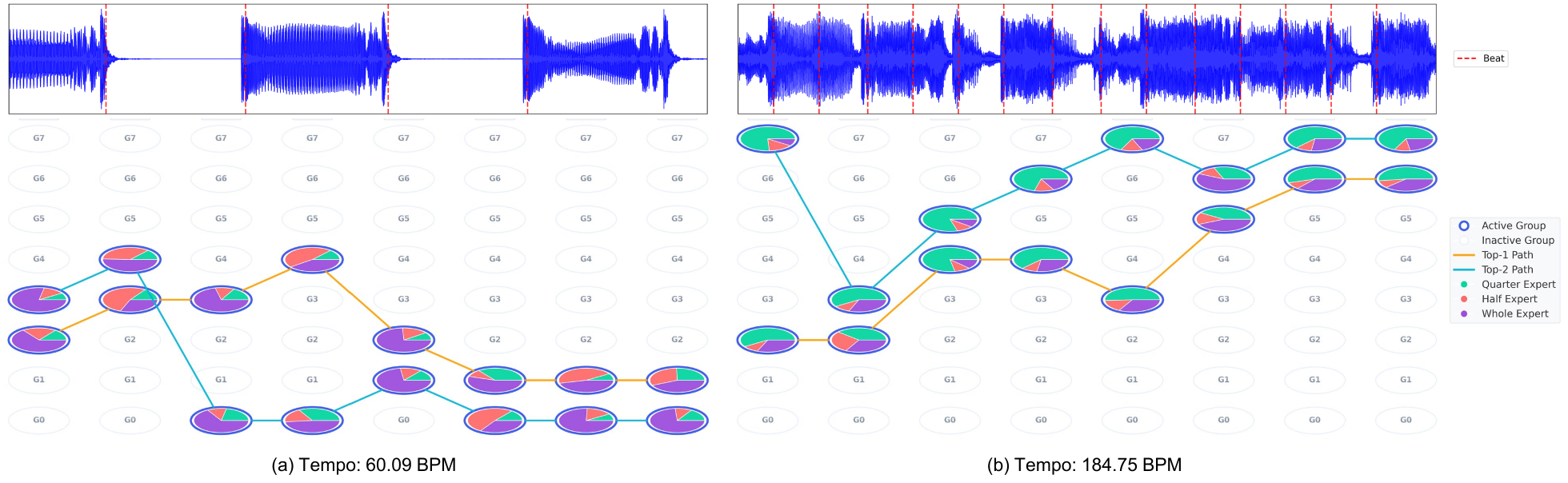}
    \vspace{-8pt}
\caption{Sample-wise routing for slow (64.09 BPM) and fast (184.75 BPM) samples. Slow tempo engages low-BPM groups and transitions from quarter- to whole-beat experts to capture long-range motions, while fast tempo activates high-BPM groups and relies on quarter-beat experts for rapid, fine-grained movements.}
    \label{fig:moe_router_case_study}
\end{figure*}

\begin{figure}[t]
    \centering
    \includegraphics[width=1\linewidth]{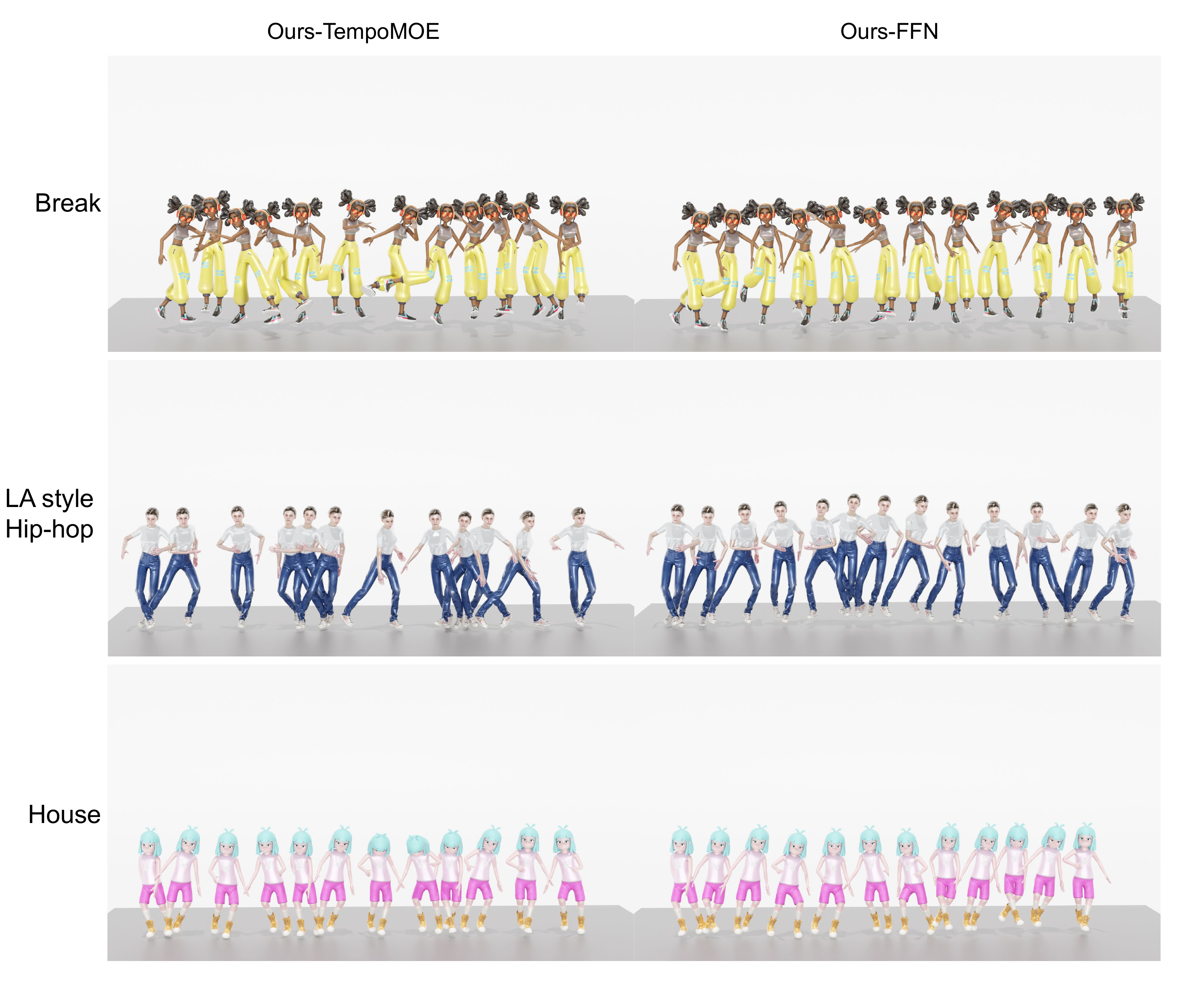}
    \vspace{-16pt}
    \caption{Qualitative comparison of TempoMoE and a standard FFN baseline across three music genres. TempoMoE produces more diverse, expressive, and rhythmically coherent motions, highlighting the advantage of tempo-aware expert routing. See supplementary videos for details.}
    \vspace{-8pt}
    \label{fig:vis_abla}
\end{figure}

\begin{table}[t]
\caption{Ablation results on routing features. Genre-based routing is excluded from PopDanceSet due to missing labels.}
\label{tab:abla_router_features_datasets}
    \centering

\renewcommand\arraystretch{1.0}
\begin{tabular}{ccccc}
     \toprule
Dataset   &   Method & ${\text{FID}_k} \downarrow$ & ${\text{Div}_k} \uparrow$  & $\text{BAS} \uparrow$ \\
     \midrule
\multirow{3}{*}{AIST++}  
&  Music & \textbf{25.13} & \textbf{7.98} & \textbf{0.2446} \\
&  Dance & 31.26 & 6.68 & 0.2136 \\
&  Genre & 29.24 & 6.92 & 0.2268 \\
\midrule
\multirow{3}{*}{FineDance}  
&  Music & \textbf{38.42} & \textbf{8.57} & \textbf{0.2316} \\
&  Dance & 44.28 & 7.12 & 0.2172 \\
&  Genre & 52.48 & 6.12 & 0.2018 \\
\midrule
\multirow{3}{*}{PopDanceSet}  
&  Music & \textbf{23.42} & \textbf{7.54} & \textbf{0.2482} \\
&  Dance & 26.48 & 6.72 & 0.2272 \\
&  Genre & -- & -- & -- \\
  \bottomrule
\end{tabular}

\end{table}

\paragraph{Model Efficiency.} Table~\ref{tab:model_efficiency} compares model architectures and runtime efficiency. Unlike some prior methods that require multiple models or high-dimensional features, such as Jukebox embeddings that need an additional model for extraction and are much slower, our approach only uses a single model with lightweight 35-dim Librosa features, which are fast to compute. Despite the reduced model complexity, it achieves competitive performance while maintaining fast inference, generating 1024 motion frames in just 0.6 seconds. This runtime measures only the motion generation and does not include feature extraction; in practical usage, including feature computation would make our efficiency advantage even more significant. This demonstrates the efficiency and practicality of our framework for real-time or large-scale applications.

\subsection{Ablation Study}

\paragraph{Expert Group.}
We investigate the impact of expert design along two axes: structural homogeneity (homogeneous vs. heterogeneous) and temporal resolution (single-scale vs. multi-scale). Specifically, we compare:
(1) Homo. Single-Scale: All groups use identical experts with a fixed 1/2-beat scale;
(2) Homo. Multi-Scale: All groups share the same set of filters at 1/4, 1/2, and 1 beat;
(3) Hetero. Multi-Scale (Ours): Each group uses tempo-specific filters adapted to its target BPM range.
As shown in Table~\ref{tab:abla_expert_group}, our design yields the best results, highlighting the benefits of intra-group scale diversity and inter-group specialization for modeling rhythmically coherent and tempo-adaptive dance.

\paragraph{Multiple Beat-Scale Experts.}
We examine whether using multiple beat scales within each expert group enhances performance. We compare four settings:
(1) Quarter-Only, (2) Half-Only, and (3) Whole-Only, where all experts use a single fixed scale;
(4) Mixed (Ours), where each group contains all three beat-scale experts.
As shown in Table~\ref{tab:abla_expert_multiple_beat_scale_expert}, the Mixed setting achieves the highest scores, confirming the benefit of modeling motion dynamics across varied rhythmic resolutions for better tempo adaptivity.

\paragraph{Number of Expert Groups.}
As shown in Table \ref{tab:abla_group_num}, using 8 expert groups achieves the best overall performance. When the number of groups is too large (e.g., 16 or 32), the BPM intervals between experts become overly small, leading to overlapping rhythmic ranges and increased training difficulty. Conversely, using too few groups (e.g., 4) limits the model’s ability to capture diverse rhythmic variations. The 8-group configuration provides a good balance, as its BPM interval (20 BPM) aligns with prior studies~\cite{dance_tempo_6_disco_time_split_20,dance_tempo_7_class} and practical dance perception thresholds where rhythm differences of about 20 BPM are perceptually distinguishable.

\begin{table}[t]
\vspace{-8pt}
\caption{Ablation results on Inter-Group Routing. Activating the top-2 adjacent tempo groups (TOP-2) achieves the best performance, demonstrating that limited hard routing with complementary group overlap improves robustness to BPM variations.}
\vspace{-4pt}
\label{tab:abla_inter_group_routing}
    \centering
\renewcommand\arraystretch{1.0}
\begin{tabular}{l|ccc}
     \toprule
     Method & ${\text{FID}_k} \downarrow$ & ${\text{Div}_k} \uparrow$  & $\text{BAS} \uparrow$ \\
     \midrule
     TOP-1 & 27.36 & 7.16 & 0.2326 \\
     TOP-2 (Ours) & \textbf{25.13} & \textbf{7.98} & \textbf{0.2446} \\
     Soft & 33.53 & 6.32 & 0.2136 \\
     Average & 31.28 & 6.12 & 0.2142 \\
     \bottomrule
\end{tabular}

\end{table}

\begin{table}[t]
\vspace{-8pt}
\caption{Ablation results on Intra-Group Routing. Soft fusion of multi-scale beat experts within each group improves motion quality and rhythmic alignment.}
\vspace{-8pt}
\label{tab:abla_intra_group_routing}

    \centering
\renewcommand\arraystretch{1.0}
\begin{tabular}{l|ccc}
     \toprule
     Method & ${\text{FID}_k} \downarrow$ & ${\text{Div}_k} \uparrow$  & $\text{BAS} \uparrow$ \\
     \midrule
     TOP-1 & 29.53 & 6.59 & 0.2176 \\
     TOP-2 & 27.32 & 7.16 & 0.2243 \\
     Soft (Ours) & \textbf{25.13} & \textbf{7.98} & \textbf{0.2446} \\
     Average & 33.16 & 6.12 & 0.2156 \\
     \bottomrule
\end{tabular}

\end{table}

\paragraph{Routing Feature.}
As shown in Table~\ref{tab:abla_router_features_datasets}, we evaluate three types of features for expert routing: 
(1) Dance: motion features; 
(2) Genre: genre embedding features; 
(3) Music (Ours): music features.
Genre routing performs poorly on FineDance, which contains 19 long-tailed sub-genres with highly imbalanced distributions. Dance routing underperforms particularly on PopDanceSet, where motion sequences often suffer from noise. In contrast, music features provide stable and anticipatory signals, enabling more accurate and generalizable routing across datasets.

\paragraph{Inter-Group Routing.}
To assess expert group selection strategies, we compare:
(1) TOP-1: activate only the top-1;
(2) TOP-2 (Ours): activate the top-2;
(3) Soft: compute a weighted sum over all groups;
(4) Average: assign equal weight to all groups.
Table~\ref{tab:abla_inter_group_routing} shows that TOP-2 performs best. Activating two adjacent groups improves robustness to BPM variations and avoids loss of specialization observed when using soft or uniform routing. This confirms the value of limited hard routing with complementary group overlap.

\paragraph{Intra-Group Routing.}
We explore fusion strategies within each group:
(1) TOP-1, (2) TOP-2, (3) Soft (Ours), and (4) Average.
As shown in Table~\ref{tab:abla_intra_group_routing}, the soft routing achieves the best performance, as it enables fine-grained rhythmic control by adaptively integrating quarter-, half-, and whole-beat experts. 
In contrast, hard selection (TOP-1/2) and uniform averaging may lead to information loss.

\begin{table}
\vspace{-8pt}
\caption{Ablation results on the number of Tempo Expert Groups. Using 8 groups achieves the best balance between capturing diverse rhythmic variations and maintaining training stability, as the 20-BPM interval aligns with human perceptual thresholds.}
\vspace{-8pt}
\label{tab:abla_group_num}
    \centering
\renewcommand\arraystretch{1.0}
\begin{tabular}{l|ccc}
     \toprule
     Groups & ${\text{FID}_k} \downarrow$ & ${\text{Div}_k} \uparrow$  & $\text{BAS} \uparrow$ \\
     \midrule
     4 & 27.64 & 7.08 & 0.2328 \\
     8 & \textbf{25.13} & \textbf{7.98} & \textbf{0.2446} \\
     16 & 33.53 & 6.32 & 0.2146 \\
     32 & 38.28 & 6.03 & 0.2092 \\
     \bottomrule
\end{tabular}

\end{table}

\paragraph{Visual Comparison with FFN Baseline.}
We replace the FFN in the baseline with our TempoMoE to evaluate the impact of expert routing. As shown in Figure~\ref{fig:vis_abla}, the FFN baseline produces repetitive, less expressive motions and struggles to capture rhythmic changes or genre-specific patterns. In contrast, TempoMoE generates fluid, diverse, and rhythmically synchronized sequences that better reflect both musical structure and stylistic nuances. This demonstrates that Tempo-Structured Expert Groups and Rhythm-Guided Routing effectively leverage tempo cues to model temporal dynamics more accurately than a standard FFN.

\subsection{TempoMOE and Routing Analysis}

\paragraph{Sample-Level Dynamics.}
Figure~\ref{fig:moe_router_case_study} illustrates routing behavior for two representative samples: one slow (64.09 BPM) and one fast (184.75 BPM). In the slow-tempo case, the model gradually shifts from quarter- to whole-beat experts and activates low-BPM expert groups, capturing longer-range structure. In contrast, the fast-tempo case consistently selects high-BPM groups and relies more heavily on quarter-beat experts to model rapid, fine-grained motion. These patterns demonstrate the router’s ability to dynamically adapt routing strategies to the input tempo. 
We further perform a dataset-level statistical analysis of the routing behavior across multiple datasets in Appendix~\ref{sec:moe_statis_router_dataset_statis}.

\noindent \textbf{See supplementary material for more results and videos.}

\section{Conclusion}
We present TempoMoE, a tempo-aware dance generation framework that integrates Tempo-Structured Expert Groups with Rhythm-Guided Routing to synthesize beat-aligned and expressive 3D dance motion sequences. By hierarchically aligning generation with rhythmic structures and dynamically routing inputs to tempo-specialized experts, our method effectively adapts to diverse tempos and complex musical patterns. Extensive experimental results confirm that TempoMoE achieves strong performance in both dance motion quality and generation efficiency.

{
    \small
    \bibliographystyle{ieeenat_fullname}
    \bibliography{main}
}
\clearpage
\appendix
\setcounter{page}{1}
\maketitlesupplementary

\begin{figure*}[t]
    \centering
    \includegraphics[width=0.95\linewidth]{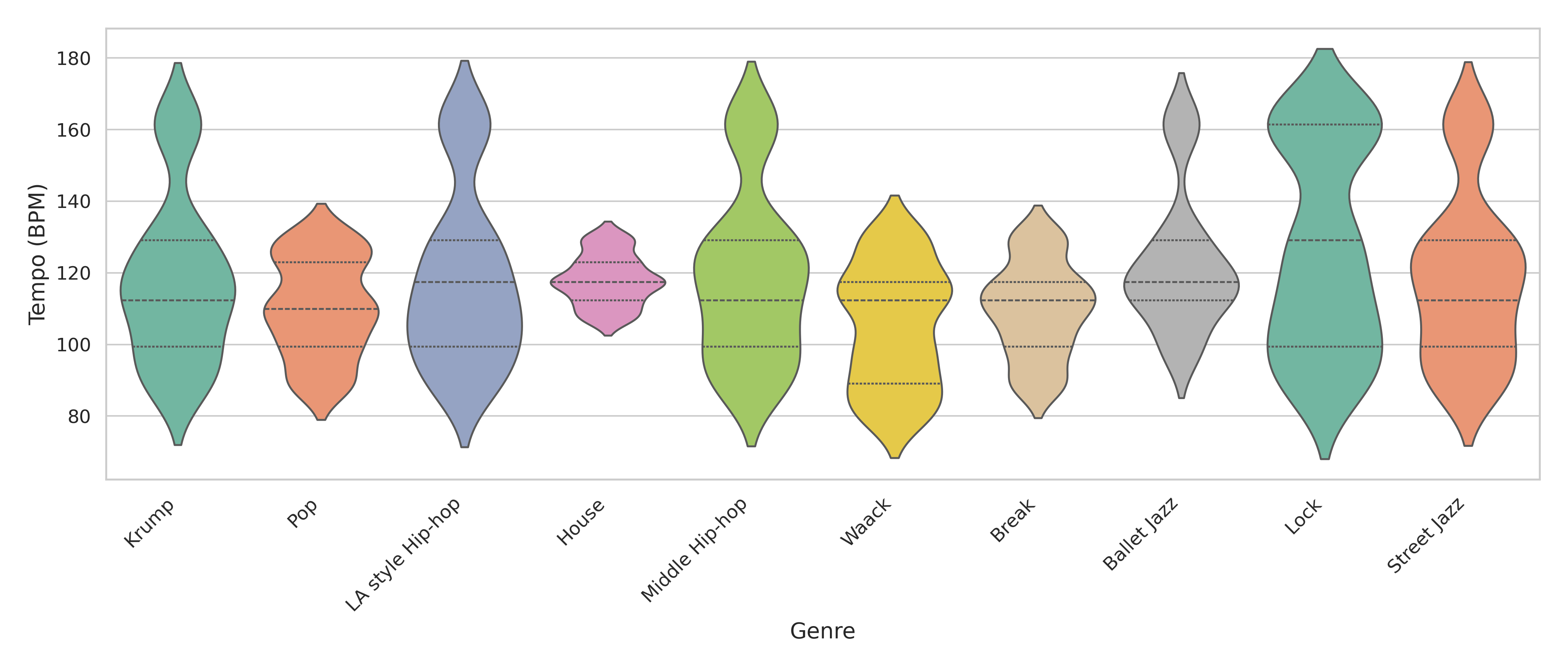}
    \vspace{-14pt}
    \caption{Tempo distributions across different genres in AIST++, visualized with violin plots. 
    Despite being grouped under the same genre, tempo values vary widely—typically between 60 and 200 BPM—indicating substantial rhythmic diversity within each genre. 
    This motivates modeling rhythmic patterns based on tempo rather than coarse or ambiguous genre labels.
 } 
    \vspace{-12pt}
    \label{fig:app_per_genre_tempo_distribution_aistpp}
\end{figure*}

\begin{figure*}
    \centering
    \includegraphics[width=0.95\linewidth]{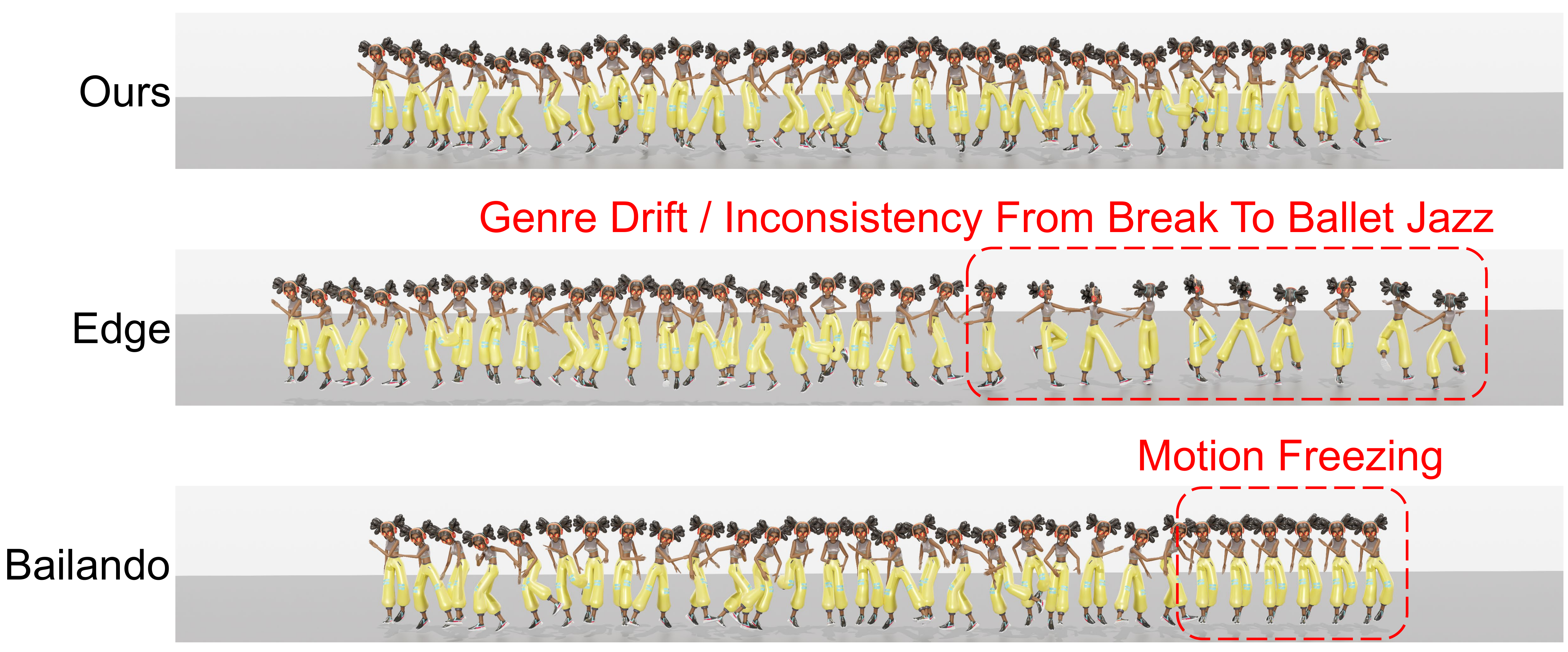}
    \vspace{-8pt}
\caption{Qualitative comparison on the Break genre. The red boxes highlight various motion issues, including motion freezing, genre drift, and penetration artifacts. Our method maintains consistent style and motion quality throughout the sequence.}
    \vspace{-12pt}

    \label{fig:app_vis_example_break}
\end{figure*}

\section{Appendix}
\subsection{ Details of Per-Genre Tempo Distribution} 

To further examine the relationship between musical genre and tempo, we visualize the tempo distribution for each genre in AIST++ using violin plots. As shown in Figure~\ref{fig:app_per_genre_tempo_distribution_aistpp}, even within the same genre, tempo varies substantially—typically spanning the 60–200 BPM range. This observation highlights the high intra-genre tempo variance, suggesting that genre alone is insufficient to characterize rhythmic properties. Instead, tempo provides a more direct and fine-grained cue for rhythm-aware modeling. This further justifies our decision to build expert groups and routing mechanisms based on tempo rather than genre.

\subsection{ Details of Tempo-Structured Expert Groups}

\noindent
\textbf{Hierarchical Temporal Modeling.}
To effectively capture the multi-scale rhythmic structure inherent in music and dance, we construct {Tempo-Structured Expert Groups} that operate across a range of tempos and beat granularities. This hierarchical design ensures that each expert processes motion features aligned with musically meaningful time scales, from fast beat-level accents to slower phrase-level gestures.

\noindent
\textbf{Tempo Discretization.}
We discretize the target tempo range of $60$ to $200$ beats per minute (BPM) into eight overlapping tempo bands, each centered at an anchor value:
\begin{equation}
{60, 80, 100, 120, 140, 160, 180, 200} \text{ BPM}.
\end{equation}
Each anchor corresponds to a dedicated expert group responsible for processing motions that align with the rhythmic dynamics of that tempo regime.

\noindent
\textbf{Beat-Scale Expert Hierarchy.}
Within each tempo group, we define three experts aligned with different rhythmic granularities:
\begin{itemize}
\item \textbf{Quarter Expert}: $\frac{1}{4}$-beat resolution
\item \textbf{Half Expert}: $\frac{1}{2}$-beat resolution
\item \textbf{Whole Expert}: 1-beat resolution
\end{itemize}
These experts are instantiated as 1D depthwise temporal convolutions, where the kernel size is rhythm-adaptive and computed to match the duration of each rhythmic unit.

\noindent
\textbf{Kernel Size Computation.}
Given a video frame rate $f$ (in frames per second), a target tempo $\text{BPM}$, and a beat resolution $r \in {\frac{1}{4}, \frac{1}{2}, 1}$, we first compute the number of frames per beat:
\begin{equation}
F_b = \frac{60 \times f}{\text{BPM}}.
\end{equation}
The convolutional kernel size $k$ is then given by:
\begin{equation}
k = \left\lceil r \times F_b \right\rceil_{\text{odd}},
\end{equation}
where $\left\lceil \cdot \right\rceil_{\text{odd}}$ denotes rounding up to the nearest odd integer to ensure a symmetric receptive field.

\noindent
\textbf{Example.}
Assume a standard video frame rate of $f = 30$ FPS and a tempo of $\text{BPM} = 120$. Then the number of frames per beat is:
\begin{equation}
F_b = \frac{60 \times 30}{120} = 15.
\end{equation}
For each beat resolution, the kernel size becomes:
\begin{align}
k_{\frac{1}{4}} &= \left\lceil \frac{1}{4} \times 15 \right\rceil_{\text{odd}} = \left\lceil 3.75 \right\rceil_{\text{odd}} = 5, \\
k_{\frac{1}{2}} &= \left\lceil \frac{1}{2} \times 15 \right\rceil_{\text{odd}} = \left\lceil 7.5 \right\rceil_{\text{odd}} = 9, \\
k_1 &= \left\lceil 1 \times 15 \right\rceil_{\text{odd}} = \left\lceil 15 \right\rceil = 15.
\end{align}
This ensures that the expert kernels are aligned with rhythm at the temporal resolution dictated by the music.

\noindent
\textbf{Summary.}
The kernel sizes for all tempo groups are summarized in Table~\ref{tab:expert_kernel_sizes}. This rhythm-aware construction offers three major benefits: (1) {tempo adaptivity}, aligning each expert's receptive field with music tempo; (2) {multi-scale modeling}, by incorporating beat subdivisions within each group; and (3) {musical inductive bias}, integrating tempo-aligned temporal priors into the model architecture.

\begin{table}[ht]
\caption{Kernel sizes for each expert in the Tempo-Structured Expert Groups.}
\vspace{-10pt}
\label{tab:expert_kernel_sizes}

\centering
 \resizebox{0.98\linewidth}{!}{
\begin{tabular}{cccc}
\toprule
\textbf{Tempo (BPM)} & \textbf{Quarter Expert} & \textbf{Half Expert} & \textbf{Whole Expert} \\
\midrule
60   & 9  & 15 & 31 \\
80   & 7  & 11 & 23 \\
100  & 5  & 9  & 19 \\
120  & 5  & 9  & 15 \\
140  & 5  & 7  & 13 \\
160  & 3  & 7  & 11 \\
180  & 3  & 5  & 11 \\
200  & 3  & 5  & 9  \\
\bottomrule
\end{tabular}}

\end{table}

\subsection{Detailed Comparison of Model Efficiency}
All experiments were conducted on the same machine equipped with an NVIDIA A6000 GPU. The inference time reported in Table~\ref{tab:app_efficiency_comparison} reflects the average runtime required to generate 1024 frames of long-term dance motion, excluding the time for music feature extraction. Notably, methods relying on high-dimensional audio representations such as Jukebox incur substantially higher preprocessing costs compared to approaches using lightweight Librosa features. As a result, their actual end-to-end latency would be even higher in practical deployment.

In addition to faster runtime, our model is designed to be independent of input sequence length during inference. In contrast, existing baselines are typically constrained by fixed-length training inputs and tend to generate sequences with matching durations. Furthermore, our architecture does not require input truncation or padding, enhancing robustness and practicality for real-world use.

Moreover, our method achieves significant efficiency gains by employing a single model and requiring only 10 denoising steps for dance motion generation. This design yields a $2\times$–$5\times$ inference speedup, while simultaneously producing diverse higher-quality and more rhythmically aligned dance sequences. In contrast, previous approaches typically rely on 50–100 iterative steps and, in some cases, multiple model components. 

\begin{table}[ht]
\vspace{-4pt}
\caption{Comparison of model architecture, music features, and inference efficiency. Inference time excludes music feature extraction. Our method achieves significant efficiency gains by employing a single model and requiring only 10 denoising steps for motion generation, yielding a $2\times$–$5\times$ inference speedup.}
\vspace{-4pt}

\label{tab:app_efficiency_comparison}
\centering
 \resizebox{0.98\linewidth}{!}{
\begin{tabular}{lcccccc}
\toprule
\textbf{Method} & \textbf{Music Feature} & \textbf{\#Model} & \textbf{\#Params (M)} & \textbf{Steps} & \textbf{Time (s)} \\
\midrule
EDGE   & Jukebox 4800 dim & Single & 49.91  & 50  & 1.3  \\
POPDG  & Jukebox 4800 dim & Single & 101.41  & 50  & 2.8  \\
Lodge  & Librosa 35 dim   & Two    & 108.23 & 100 & 0.8  \\
\textbf{Ours}  & Librosa 35 dim   & Single & 70.13  & 10  & 0.6 \\
\bottomrule
\end{tabular}}

\end{table}

\subsection{Detailed Example of the Break Genre Dance}

As shown in Figure~\ref{fig:app_vis_example_break}, we present a representative example of a generated Break dance sequence to illustrate qualitative differences across models.

EDGE demonstrates a genre inconsistency issue: although the generated motion initially exhibits breakdance-like characteristics, it gradually drifts toward movements more reminiscent of ballet jazz. This degradation is primarily due to EDGE’s limited rhythmic sensitivity—its generation process lacks explicit modeling of beat alignment or tempo variation, making it prone to producing stylistically incompatible segments, especially in rhythm-intensive genres like Break. The red boxes highlight abrupt stylistic shifts and rhythm-misaligned transitions that arise as a result.

Bailando, on the other hand, suffers from motion stalling and degraded dynamics in the latter part of the sequence. This issue stems from cumulative decoding errors and the rigidity of its frozen VQ-VAE motion representations. As generation proceeds autoregressively, small prediction inaccuracies compound over time, and the fixed token space limits the model’s flexibility to correct or adapt to evolving musical cues. Consequently, the decoder struggles to sustain diverse and context-aware motion patterns, leading to reduced diversity and repetitive motion toward the end.

In contrast, our method maintains strong rhythmic awareness and stylistic consistency throughout the sequence. By leveraging tempo-aware experts and routing, it produces fluid breakdance motions that remain well-synchronized with the underlying music.

\begin{figure*}[t]
    \centering
    \includegraphics[width=0.99\linewidth]{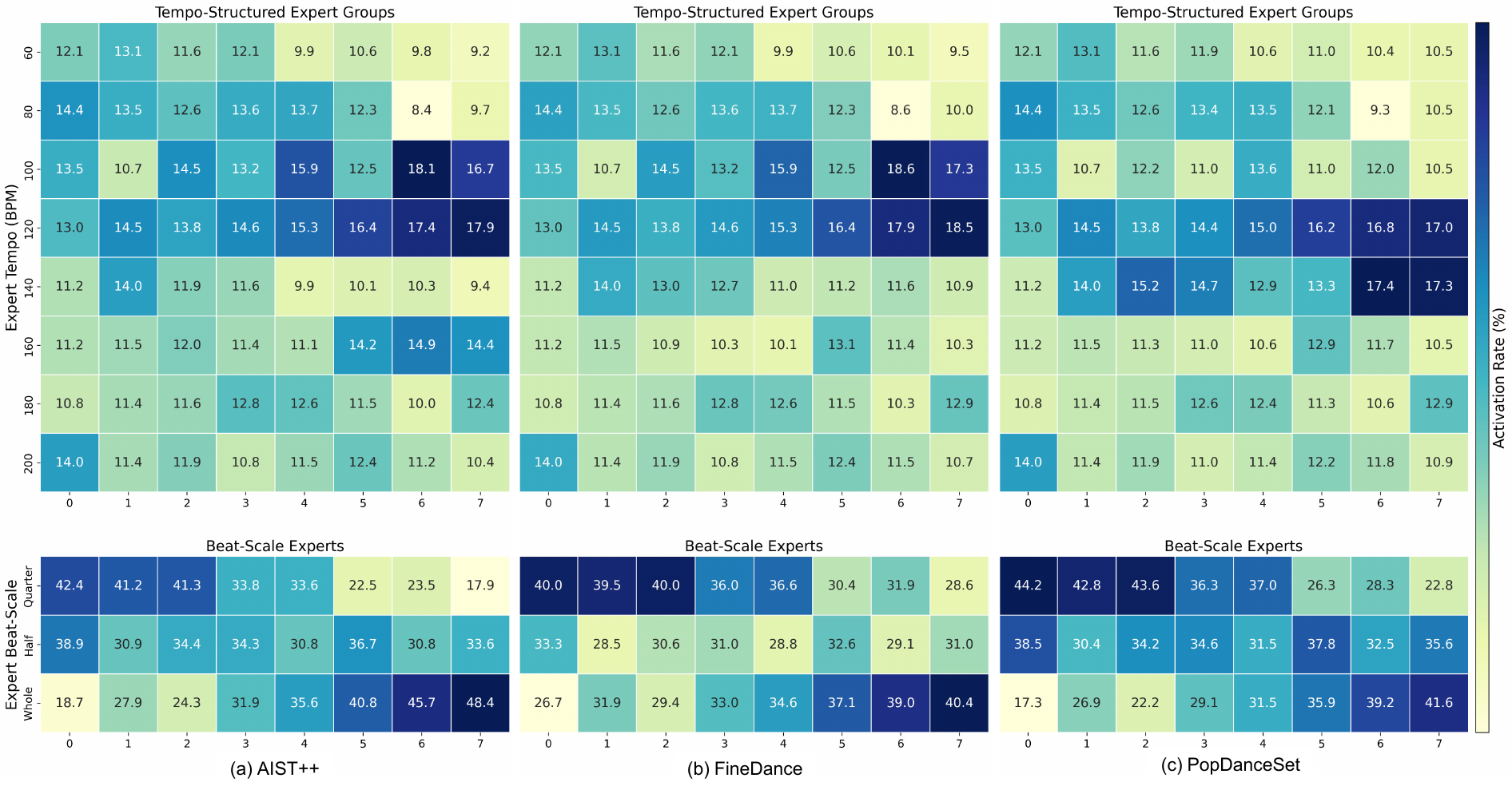}
\caption{Dataset-level routing statistics across AIST++, PopDanceSet, and FineDance. TempoMoE exhibits distinct BPM and beat-scale activations for each dataset: AIST++ peaks near 100 and 160 BPM, PopDanceSet centers around 120–140 BPM with a preference for quarter-beat experts, and FineDance shows broader BPM coverage with greater reliance on whole-beat experts. Across all datasets, lower layers capture fine-grained rhythmic details, while higher layers focus on long-term motion structure. These patterns demonstrate that TempoMoE effectively adapts to diverse tempo distributions and hierarchical rhythmic structures, supporting rhythmically accurate and expressive dance generation.}
    \label{fig:moe_statis_router_dataset_statis}
\end{figure*}

\subsection{ Visualizing Motion Differences across Tempos}
\label{sec:tempo_effect}

\begin{figure*}
    \centering
    \includegraphics[width=1\linewidth]{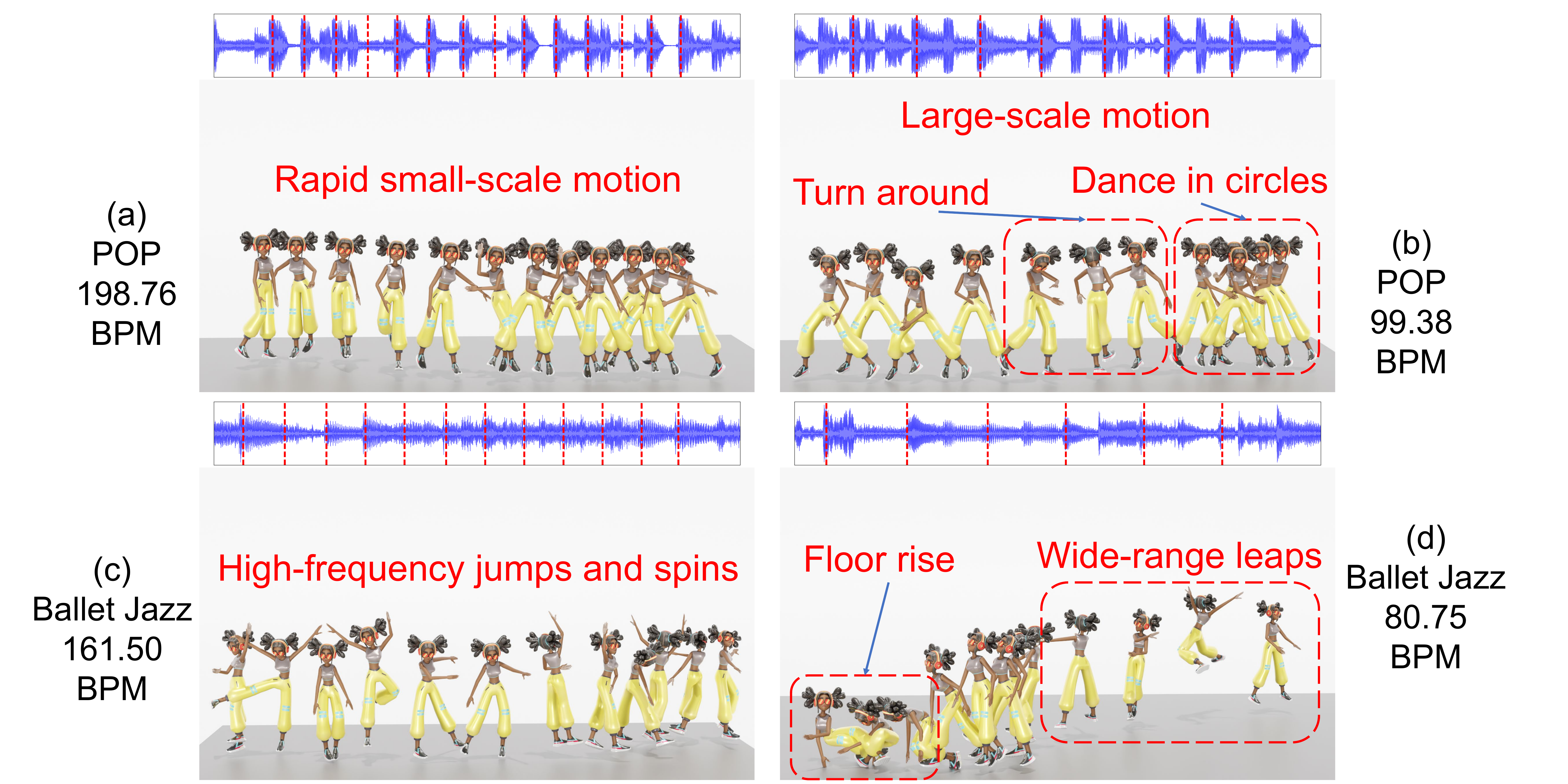}
    \caption{Qualitative visualization of generated dances under different tempo settings across two music genres. (a) Pop, 198 BPM: Fast tempo induces dense and localized motions, including quick arm swings and high-frequency isolations. (b) Pop, 99 BPM: Slower tempo allows broader gestures, such as body turns and full-body transitions. (c) Jazz Ballet, 160 BPM: High tempo yields compact yet energetic movements, featuring frequent spins and small vertical jumps. (d) Jazz Ballet, 80 BPM: Low tempo enables expansive, expressive choreography with floor-to-stand transitions and wide-range leaps. These results demonstrate our model's ability to disentangle tempo and genre for controllable and style-consistent dance generation.}
    \label{fig:app_vis_tempo}
\end{figure*}

Figure~\ref{fig:app_vis_tempo} presents qualitative comparisons of generated dance on two music genres—Pop and Jazz Ballet—each rendered under two distinct tempo settings: fast (198 or 160 BPM) and slow (99 or 80 BPM), denoted as (a)-(d). This analysis isolates the effect of tempo modulation while holding genre constant, allowing us to examine how the model adjusts motion patterns in response to rhythmic pacing.

For Pop dance in (a) and (b), we observe clear differences in both motion pacing and spatial extent. At 198 BPM (a), the generated sequence is characterized by high-frequency, localized actions such as rapid arm swings, quick foot taps, and upper-body isolations. The choreography aligns tightly with the fast-paced beat, resulting in compact and rhythm-synchronized motions. At 99 BPM (b), the slower rhythm enables more expansive and fluid movements, including body turns, full-body transitions, and broader steps. This demonstrates the model’s ability to modulate motion granularity and articulation based on tempo, while preserving the stylistic traits of Pop dance.

For Jazz Ballet in (c) and (d), tempo adaptation manifests differently due to genre-specific characteristics. At 160 BPM (c), the model generates tight vertical jumps, swift torso movements, and frequent spins, reflecting the genre’s dynamic yet refined motion vocabulary. In contrast, the 80 BPM version (d) exhibits low-to-high transitions (e.g., rising from the floor), sweeping arm gestures, and wide-range leaps. These motions reflect greater spatial extension and structural expressiveness, enabled by the slower beat. Notably, the model maintains genre-appropriate elegance while adapting motion density and timing.

Overall, these results confirm that our method enables precise and expressive choreography control across diverse music conditions.

\subsection{Dataset-Level Statistics of Routing Analysis.}
\label{sec:moe_statis_router_dataset_statis}

Figure~\ref{fig:moe_statis_router_dataset_statis} summarizes aggregated routing across datasets. AIST++ shows prominent activations near 100 and 160 BPM; PopDanceSet concentrates around 120–140 BPM and favors quarter-beat experts; FineDance displays a broader BPM distribution with greater reliance on whole-beat experts, reflecting its emphasis on expressive, phrase-level choreography. Across all datasets, lower layers prioritize finer rhythmic details, while higher layers increasingly attend to long-term structure—indicating effective hierarchical rhythm modeling.

\subsection{Visual Comparison with FFN Baseline}
\label{sec:abla_ffn}

\begin{figure*}[t]
    \centering
    \includegraphics[width=1\linewidth]{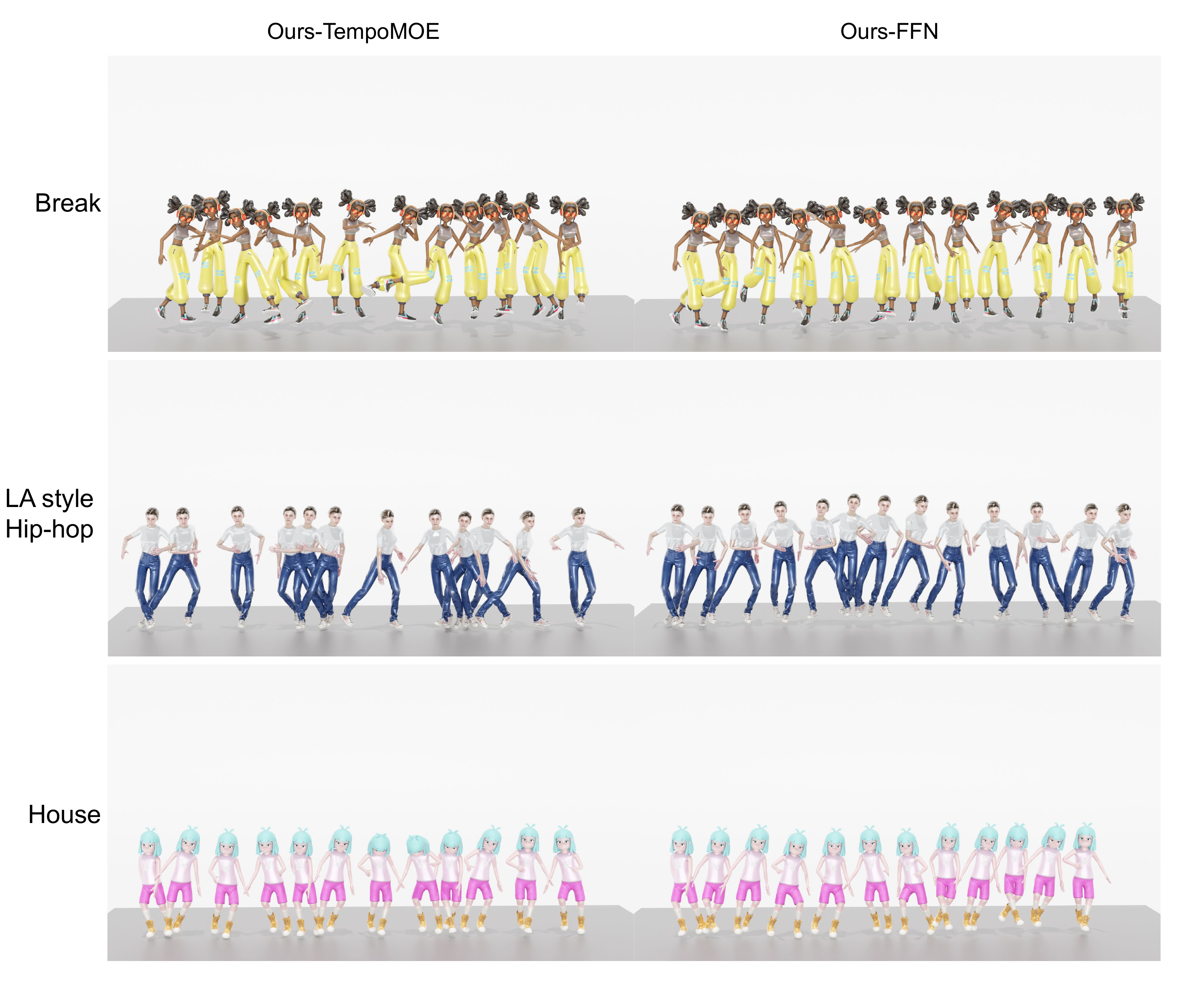}
    \caption{Qualitative comparison between our TempoMoE model and a standard FFN-based baseline across three music genres. TempoMoE generates more diverse, expressive, and rhythmically coherent motions that align better with musical phrasing and stylistic priors. This highlights the advantage of our tempo-aware mixture-of-experts architecture in capturing genre-specific and rhythm-sensitive choreography patterns. See supplementary videos for details.}
    \label{fig:app_vis_abla}
\end{figure*}

We evaluate TempoMoE by comparing it to a baseline using FFN instead of expert routing. As shown in Figure~\ref{fig:app_vis_abla}, the FFN baseline tends to produce repetitive, less expressive motions and often fails to accurately align with rhythmic changes or genre-specific characteristics. In contrast, replacing the FFN with TempoMoE leads to motions that are significantly more fluid, diverse, and rhythmically synchronized with the input music. TempoMoE sequences faithfully capture both fine-grained musical structure, such as beat-level accents, and broader stylistic nuances, including genre-adapted motion patterns. These comparisons clearly demonstrate the effectiveness of Tempo-Structured Expert Groups and Hierarchical Rhythm-Guided Routing. By dynamically leveraging tempo cues to route different motion primitives through specialized experts, TempoMoE can model temporal dynamics more accurately and flexibly than a flat FFN baseline, producing higher-quality and more musically coherent dance motions.

\subsection{Failure Cases on Latest Out-of-Distribution TikTok Samples}

The supplementary materials include several test cases using trending TikTok dance music. These clips are often remix-heavy and user-curated, featuring irregular rhythms, abrupt transitions, and non-canonical beat structures~\cite{m2d_tiktok_new,m2d_tiktok_dataset,m2d_tiktok_analysis,m2d_tiktok_analysis_2}, which pose significant challenges for music-to-dance generation and highlight the need for more robust modeling strategies in real-world, user-generated content scenarios.   As shown in the demo videos, both EDGE and our method struggle under these conditions. EDGE, which relies on high-level Jukebox embeddings, often fails to capture meaningful rhythmic patterns, producing static or off-beat motions. Our method performs comparatively better due to the use of low-level Librosa features combined with tempo-aware routing, which maintain more stable distributions across diverse music inputs. This allows for more consistent rhythm perception and slightly improved motion

\end{document}